\documentclass[lettersize,journal]{IEEEtran}
\usepackage{amsmath,amsfonts}
\usepackage{algorithmic}
\usepackage{algorithm}
\usepackage{array}
\usepackage[caption=false,font=normalsize,labelfont=sf,textfont=sf]{subfig}
\usepackage{textcomp}
\usepackage{stfloats}
\usepackage{url}
\usepackage{verbatim}
\usepackage{graphicx}
\usepackage{cite}
\usepackage{multirow}
\usepackage[table,xcdraw]{xcolor}
\usepackage[colorlinks,linkcolor=red]{hyperref}

\usepackage{arydshln}

\begin{document}

\title{Boosting Adversarial Transferability with Learnable Patch-wise Masks}

\author{Xingxing~Wei$^*$~\IEEEmembership{Member,~IEEE}, and Shiji~Zhao
        % <-this % stops a space
\thanks{Xingxing Wei, Shiji Zhao were at the Institute of Artificial Intelligence, Beihang University, No.37, Xueyuan Road, Haidian District, Beijing,
100191, P.R. China. (E-mail: \{xxwei, zhaoshiji123\}@buaa.edu.cn).}% <-this % stops a space
\thanks{Xingxing Wei is the corresponding author.}

}

\markboth{IEEE Transactions on Multimedia}%
{Shell \MakeLowercase{\textit{et al.}}: A Sample Article Using IEEEtran.cls for IEEE Journals}


\maketitle

\begin{abstract}
Adversarial examples have attracted widespread attention in security-critical applications because of their transferability across different models. Although many methods have been proposed to boost adversarial transferability, a gap still exists between capabilities and practical demand. In this paper, we argue that the model-specific discriminative regions are a key factor causing overfitting to the source model, and thus reducing the transferability to the target model. For that, a patch-wise mask is utilized to prune the model-specific regions when calculating adversarial perturbations. To accurately localize these regions, we present a learnable approach to automatically optimize the mask. Specifically, we simulate the target models in our framework, and adjust the patch-wise mask according to the feedback of the simulated models. To improve the efficiency, the differential evolutionary (DE) algorithm is utilized to search for patch-wise masks for a specific image. During iterative attacks, the learned masks are applied to the image to drop out the patches related to model-specific regions, thus making the gradients more generic and improving the adversarial transferability. The proposed approach is a preprocessing method and can be integrated with existing methods to further boost the transferability. Extensive experiments on the ImageNet dataset demonstrate the effectiveness of our method. We incorporate the proposed approach with existing methods to perform ensemble attacks and achieve an average success rate of 93.01\% against seven advanced defense methods, which can effectively enhance the state-of-the-art transfer-based attack performance.
\end{abstract}

\begin{IEEEkeywords}
DNNs, Adversarial Attack, Adversarial Transferability.
\end{IEEEkeywords}

\section{Introduction}
\IEEEPARstart{D}{eep} neural networks (DNNs) have achieved remarkable performance in a variety of computer vision tasks, including image classification \cite{krizhevsky2012imagenet,simonyan2014very}, object detection \cite{girshick2015fast} and semantic segmentation \cite{long2015fully}.  However, DNNs are vulnerable to adversarial examples \cite{szegedy2013intriguing, DBLP:journals/corr/GoodfellowSS14}, which are generated by adding well-designed imperceptible perturbations to benign images, making deep models output wrong predictions expected by the adversary. More importantly, adversarial examples show transferability, which means that the adversarial examples crafted for one DNN model may cause other unseen DNN models to make mistakes. These properties make adversarial examples attract widespread attention, especially in security-sensitive application scenarios, e.g., face recognition \cite{wei2022adversarial,wei2022simultaneously}, self-driving \cite{liu2019perceptual}.

\begin{figure*}
	\centering
	\includegraphics[width=1\linewidth]{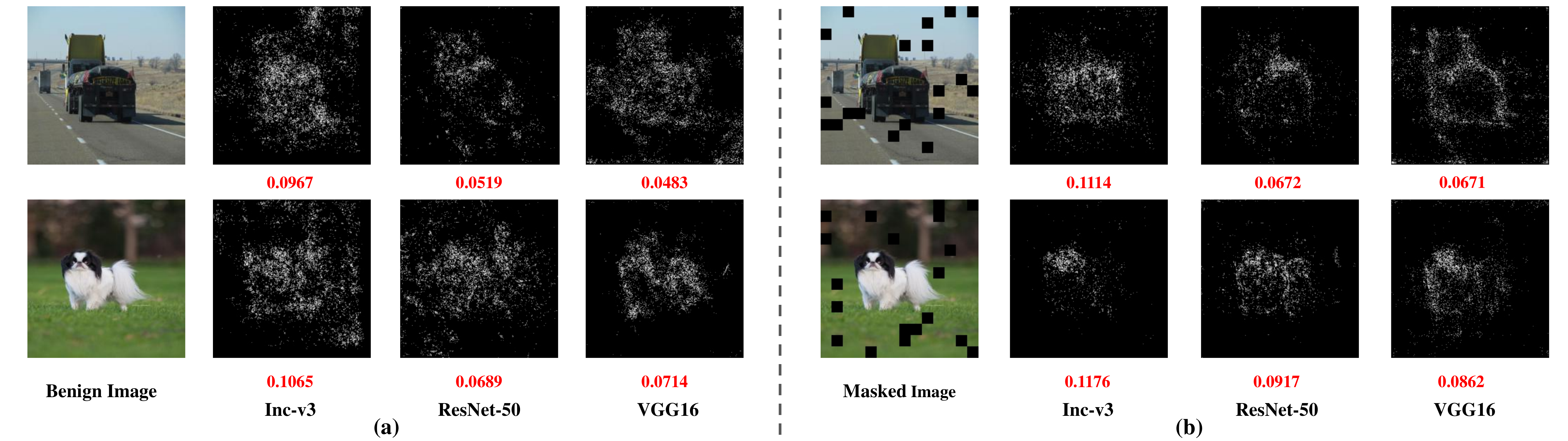}
	\caption{Illustrations of the discriminative regions. We adapt smoothed gradient \cite{smilkov2017smoothgrad} to visualize the saliency maps of three DNN models, i.e., ResNet=50 \cite{he2016deep}, Inception-v3 \cite{szegedy2016rethinking} and VGG-16 \cite{simonyan2014very}. The left denotes the discriminative regions on the original images, and the right denotes the discriminative regions on the masked images. Red numbers represent the clustering coefficient \cite{watts1998collective,holland1971transitivity} of the saliency maps.}
    \label{fig:1}
\end{figure*}

Intuitively speaking, transfer-based attacks utilize the assumption that different DNN models trained on the same dataset have similar discriminative regions. Therefore, an adversarial example generated against the white-box source models can be transferred to attack the black-box target model. However, because the generated adversarial perturbation is highly correlated with the discriminative regions of the white-box source model at the given input point, overfitting may occur. We argue that the discriminative regions of a DNN model can be divided into model-generic regions and model-specific regions. As their names imply,  model-generic regions are the most similar discriminative regions across DNN models, and model-specific regions are the most differential regions for various DNN models, which leads to  overfitting and restricts transferability. For that, we illustrate the discriminative regions of three popular models by visualizing their gradients\footnote{Gradients are directly correlated with the model's discriminative regions. Here we choose a popular gradient visualization method: smoothgrad \cite{smilkov2017smoothgrad} to remove the noise effects of gradients. Other popular visualization methods can also be used.} in Figure \ref{fig:1} (a). We can see that these three models focus on similar discriminative regions, but various model-specific discriminative regions indeed exist for each DNN model, such as some highlighted saliency maps appearing in the background. When the generated adversarial perturbations overfit to the source model, these model-specific perturbations become key for mitigating the improvements versus the transferability.

From this viewpoint, a variety of existing methods can be regarded to boost transferability by searching for the model-generic discriminative regions via various strategies and then generating the transferable adversarial examples. For example,  input transformation attacks \cite{dong2019evading, xie2019improving, wu2021improving, wang2021admix, long2022frequency} adapt various input transformations to create diverse input patterns, and then average the gradients on these patterns to avoid overfitting, which actually searches for the model-generic discriminative regions via data augmentation. Feature-level object-aware attacks  \cite{wu2020boosting, wang2021feature, zhang2022improving} mitigate overfitting to specific blind spots of the source model by measuring and highlighting the important features of the object region to search for the model-generic regions.

In addition to searching for the model-generic discriminative regions, we argue that another available way exists to boost the transferability in the opposite direction, i.e., localizing and pruning these model-specific regions, and thus the complete model-generic regions can be better retained. 
However, model-specific regions change with different DNN models, and we cannot remove these regions one by one. Moreover, how to localize these model-specific regions is a challenging problem because we do not have a quantitative metric to accurately distinguish and pick the model-specific regions.

\begin{figure*}[t]
	\centering
	\includegraphics[width=0.95\linewidth]{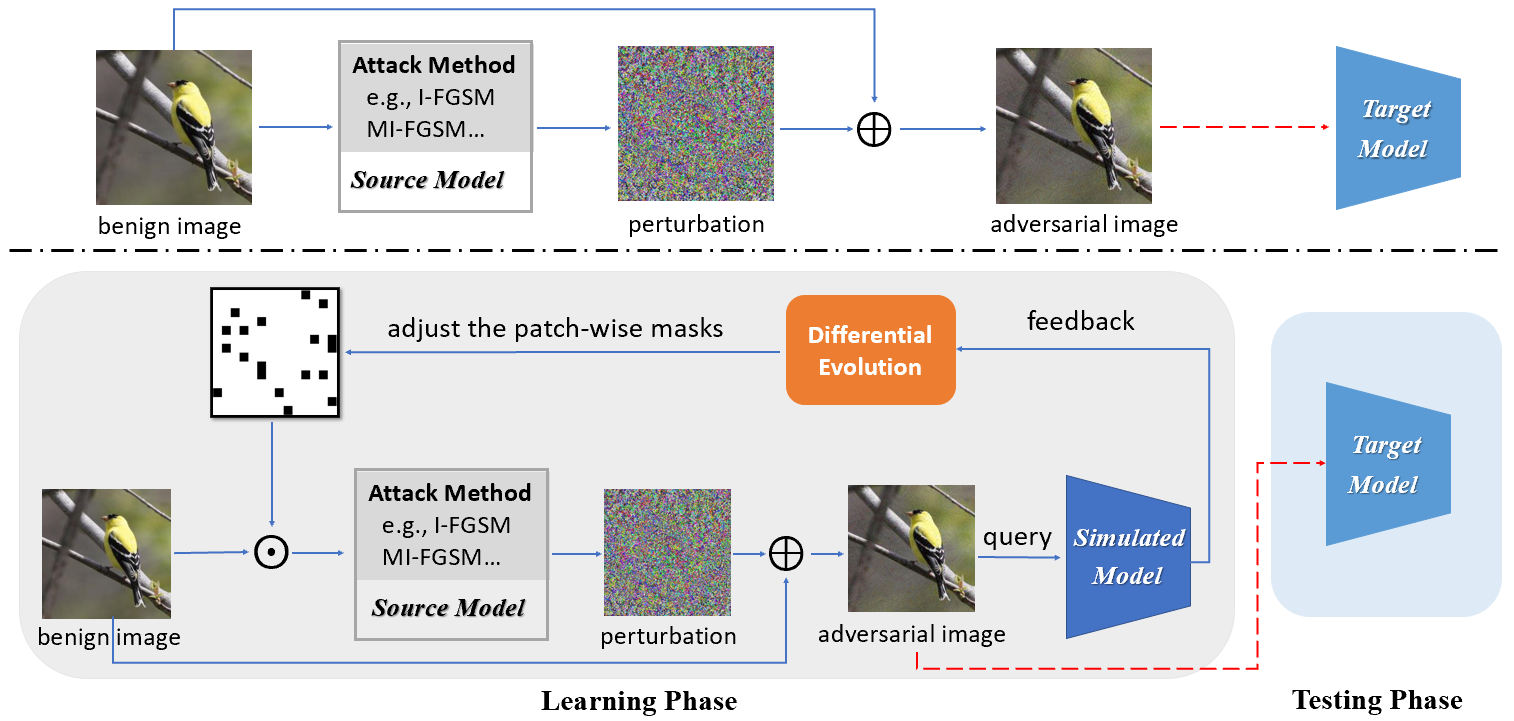}
	\caption{Pipeline of existing transfer-based attack and the proposed attack. The existing attack is illustrated above the dotted line, and the proposed learning-based attack is below the dotted line. In the learning phase, we utilize the differential evolution algorithm to learn a patch-wise mask for the specific image, which can drop out the patches related to model-specific regions and improve adversarial transferability. We feed the image into target models in the testing phase to conduct transfer-based attacks. $\odot$ is the element-wise product.}
    \label{fig:2}
\end{figure*}

To address this issue, in this paper,  we propose a preprocessing method named Learnable Patch-wise Mask  (LPM) to accurately localize the model-specific regions, and then prune them according to the learned masks.  Because discriminative regions of DNN models are based on the image's content, we aim to drop out the image's patches related to the model-specific discriminative regions, thus making the semantic information of dropped patches meaningless. In this way, different DNN models will not pay attention to model-specific regions. To accurately localize these patches, we align a patch-wise binary mask to the input image. To better learn this patch-wise mask,  we construct ensemble DNN models to simulate a target model (called simulated models). Then, we adjust the patch-wise mask according to the feedback from simulated models. If the mask is appropriate (i.e., the model-specific regions are pruned well), the transfer-based attack performance versus these simulated models will be high, and vice versa. To improve the solving efficiency, we utilize the differential evolution (DE) algorithm \cite{chakraborty2008advances} to optimize the patch-wise mask. The pipeline of our method is illustrated in Figure \ref{fig:2}. The code can be found in \url{https://github.com/zhaoshiji123/LPM}.

Because of the absent model-specific regions, the poor transferability caused by overfitting will be mitigated. The masked images and the gradient maps are illustrated in Figure \ref{fig:1} (b). We see that these models focus on more generic discriminative regions of the masked images than the benign ones. In this way, our method can be considered as a preprocessing method to generate masked images, which can be applied to enhance the performance of transfer-based attack methods when generating adversarial examples.

The main contributions can be summarized as follows:
\begin{itemize}

\item 
We argue that the model-specific discriminative regions are a key factor causing overfitting to the source model, and thus reducing the transferability. For that, we propose a preprocessing method named Learnable Patch-wise Mask (LPM) to learn and prune the model-specific regions for different DNN models in a unified manner. In this way, different models have consistent model-generic regions, and the adversarial perturbations can have better attack transferability. 

\item
We propose a learnable method that utilizes the differential evolutionary algorithm to search for a patch-wise binary mask. We use the learned masks to drop out the image patches related to the model-specific regions and boost transfer-based attacks by mitigating overfitting to the source model.

\item Extensive experimental results show that the proposed method can remarkably enhance the attack transferability of existing transfer-based attack methods in both mainstream undefended and defended models, which shows the effectiveness of our LPM.
\end{itemize}

 \par The rest of this paper is structured as follows: Section 2 reviews the related works on transfer-based adversarial attacks, and the corresponding defense methods. Section 3 describes the detailed method for the proposed method. Section 4 presents and analyzes the experimental results. Section 5 concludes the paper.
 
\section{Related Work}

\subsection{Transfer-based Adversarial Attacks}

Existing transfer-based adversarial attacks can be divided
into three classes, i.e. gradient optimization attacks \cite{dong2018boosting, lin2019nesterov,  Wang_2021_CVPR, xiong2022stochastic, gao2021push, wan2023average, yuan2021automa}, input transformation attacks \cite{dong2019evading, xie2019improving, wu2021improving, wang2021admix,  long2022frequency}, and feature-level object-aware attacks \cite{  wang2021feature, zhang2022improving,zhang2022transferable}.

The advanced gradient calculation methods can stabilize the update directions and escape from poor local maxima during the iterations, resulting in more transferable adversarial examples \cite{dong2018boosting}. Lin et al. \cite{lin2019nesterov} introduce the Nesterov accelerated gradient into iterative attacks to effectively look ahead and improve the transferability of adversarial examples. Wang et al. \cite{Wang_2021_CVPR} further consider the gradient variance of the previous iteration to tune the current gradient to stabilize the update direction and avoid overfitting.

Input transformation attacks adapt various input transformations to 
create diverse input patterns to avoid overfitting and improve adversarial transferability \cite{xie2019improving, dong2019evading}. Admix \cite{wang2021admix} calculates the gradient on the input image admixed with a small portion of each image to craft more transferable adversarial images. Wu et al. \cite{wu2021improving} claim that these prior data augmentation methods employ fixed transformations, which lead to inferior results. They propose exploiting an adversarial transformation network to automatically model various transformations. From another perspective, Long et al. \cite{long2022frequency} propose a spectrum transformation based on a discrete cosine transform to diversify images and improve adversarial transferability.

Feature-level object-aware attacks mitigate overfitting to the source model by measuring and highlighting the important features, which can be achieved by maximizing the distance between the adversarial image and the benign image on the feature map. Wang et al. \cite{wang2021feature} perform attacks in the intermediate layers by suppressing positive features and promoting negative features directly to enhance transferability. Zhang et al. \cite{zhang2022improving} propose a neuron attribution-based attack to conduct attacks with more accurate neuron importance estimations.

Existing attacks boost the transferability by searching for the model-generic discriminative regions via various strategies. However, we aim to prune the model-specific regions, which have a different mechanism to boost transferability. In the experiments, we enhance the attack transferability of state-of-the-art methods to verify the effectiveness of our method.

\subsection{Defense against Adversarial Attacks}

To defend against adversarial examples, several methods have been proposed, including adversarial training \cite{DBLP:journals/corr/GoodfellowSS14, zhao2022enhanced} and preprocessing \cite{DBLP:conf/iclr/XieWZRY18, liao2018defense}. Adversarial training injects adversarial examples into the training process. The adversarial training models can resist the perturbations in the gradient direction of the loss function, but the accuracy on benign samples will decrease. Preprocessing methods purify the adversarial examples by removing or destroying adversarial perturbations but reducing the accuracy of benign images. These defenses include the high-level representation denoiser (HGD) \cite{liao2018defense}, random resizing and padding (R\&P) \cite{xie2017mitigating}, randomization smoothing (RS) \cite{jia2019certified}, feature distillation (FD) \cite{Liu_2019_CVPR}, feature squeezing method: bit reduction (BIT) \cite{DBLP:conf/ndss/Xu0Q18} and a neural representation purifier (NRP) \cite{naseer2020self}.

We employ some representative state-of-the-art defense methods in the experiments to test the performance of our method and evaluate its effectiveness against defenses.

\section{Methodology}

% \subsection{Gradient-based Adversarial Attack}
\subsection{Preliminaries}
% \subsection{Problem formulation}

Given a classification network $f_{\theta}$ parameterized by $\theta$, let $x, y$ denote the benign image and its corresponding ground-truth label. The goal of the adversarial attack is to find an example $x^{adv}$ that is in the vicinity of $x$ but misclassified by the network. In most cases, we use the $L_{p}$ norm to limit the adversarial perturbations below a threshold $\epsilon$, where $p$ could be $0$, $2$, $\infty$. This can be expressed as:
\begin{equation}\label{eq:untarget}
f_{\theta }(x^{adv})\neq y, \ s.t. \ \left \| x^{adv}-x \right \|_{p}\leq \epsilon.
\end{equation}

The iterative version of FGSM (I-FGSM) \cite{DBLP:journals/corr/KurakinGB16} iteratively applies the fast gradient sign method multiple times with a small step size $\alpha$, which can be expressed as:
\begin{equation}\label{eq:i-fgsm}
x_{t+1}^{adv} = x_{t}^{adv} + \alpha \cdot sign(\bigtriangledown _{x_{t}^{adv}}J(x_{t}^{adv}, y)),
\end{equation}
where $x_{0}^{adv} = x$. $\bigtriangledown _{x_t^{adv}}J$ is the gradient of the loss function $J(\cdot)$ with respect to $x_{t}^{adv}$ and cross-entropy loss is often used. $sign(\cdot)$ is the sign function to limit perturbations conformin to the $L_{\infty}$ norm bound.

\subsection{Attack Based on Learnable Patch-wise Mask}

The adversarial example generated by existing transfer-based attacks has poor transferability due to overfitting to the source model. As previosly mentioned, we argue that the model-specific discriminative regions of different DNN models are the key factor causing this overfitting (see Figure \ref{fig:1}). To boost transferability, we propose a mask to drop out the image's patches related to these regions in the process of generating adversarial perturbations. To implement this idea, we employ a learnable strategy to predict a patch-wise mask (LPM) to localize the corresponding patches in the images. For convenience of description, we apply the LPM combined with the I-FGSM as an example to introduce our method. The update of perturbations in the LPM-I-FGSM is formalized as:
\begin{equation}\label{eq:lfgsm}
x^{adv}_{t+1} = x^{adv}_{t} + \alpha \cdot sign(\bigtriangledown _{x^{adv}_{t}}J(x^{adv}_{t} \odot M, y)),
\end{equation}
where $M \in \{0, 1\}^{m\times n}$ is the learned patch-wise binary matrix, and the patches on the image with $0$ will be dropped out. $m \times n$ are the total number of patches normally divided via the predefined patch size $p_s$.

% simulated models

We simulate the target model $t(\cdot)$ to compute feedback to adjust the mask and call it the simulated model $s(\cdot)$, which can be one or more DNN models. If the performance is strong on the simulated model, indicating that the adversarial examples have a good generalization, it will also perform well on the target model. We can choose a simulated model that is closely related to the target model according to prior knowledge. If no prior knowledge of the target model exists, we utilize the representative DNN models. In this work, we select ResNet-50 \cite{he2016deep}, VGG-16 \cite{simonyan2014very}, and DenseNet-161 \cite{huang2017densely}, and use the ensemble version as the simulated model. The effect of the simulated models will be explored in the ablation study.

The differential evolution (DE) algorithm is employed to optimize the patch-wise mask. The input includes randomly initialized masks $\mathcal{M}(0)$ and feedback $\phi$ on the simulated model. This can be expressed as:
\begin{equation}\label{eq:de_1}
M^* = D(\mathcal{M}(0), \phi),
\end{equation}
where $D(\cdot)$ is the DE algorithm. The feedback is crucial for adjusting the mask. The cross-entropy loss with the ground-truth label can reflect the classification results, and the loss value represents a greater propensity for misclassification. In addition,  since multiple simulated models are utilized  to guide the optimization of masks, we use both the mean and variance of cross-entropy loss as the feedback to prevent falling into the local optimal solution of a certain simulated model, which can be formalized as:
\begin{equation}\label{eq:de_ce}
\phi = -\sum _{i=0} ^{n_s} CE (p_{i}, y) +  \sum _{i=0} ^{n_s} \big ( CE (p_{i}, y) - \frac{1}{n_s} \sum _{j=0} ^{n_s} CE (p_{j}, y) \big)^2,
\end{equation}
where $n_s$ is the number of simulated models, and $y$ indicates the one-hot label. $p_{i}$ is the predicted probability by the $i$-th simulated model $s_i(\cdot)$, and $CE$ denotes the cross-entropy loss.

% overall framework

The whole framework of the proposed method is illustrated in Figure \ref{fig:2}. We initialize a patch-wise mask and combine it with the input image by an elementwise product. The adversarial perturbations are calculated by the I-FGSM attack on the source model $f(\cdot)$ for the modified input image. Then, the generated adversarial image is input to the simulated model $s(\cdot)$ to calculate the feedback $\phi$. The differential evolution algorithm adjusts the patch-wise masks according to the feedback until the optimal mask is achieved. Finally, we test the transferability of adversarial examples on the target model $t(\cdot)$. The details can be found in Algorithm \ref{alg:1}. 

In our method, the patch size is a hyperparameter and we explore its influence in the experimental section. Furthermore, to remove the randomness of the predicted mask solution, we predict multiple masks for one image and employ the intersection of multiple learned masks for every step of the final attack. We find that this method can further improve the transfer-based attack success rate.

\subsection{Learning the Mask by Differential Evolution}

The DE algorithm utilizes crossover strategy, mutation strategy, and selection strategy to heuristically search for the optimal solution with the guidance of feedback. Specifically, in our evolutionary algorithm, a population represents a set of solution masks. Given the population size $P$, the $k$-th generation solution $\mathcal{M}(k)$ is represented as:
\begin{equation}\label{eq:de1}
\mathcal{M}(k) = \{{M}_{i}(k) | {M}_{i}(k) \in \{0, 1\}^{m\times n}, i=1,...,P \},
\end{equation}
where $M_i(k)$ denotes the $i$-th individual mask solution in the $k$-th generation.

\begin{figure*}[t]
	\centering
	\includegraphics[width=0.95\linewidth]{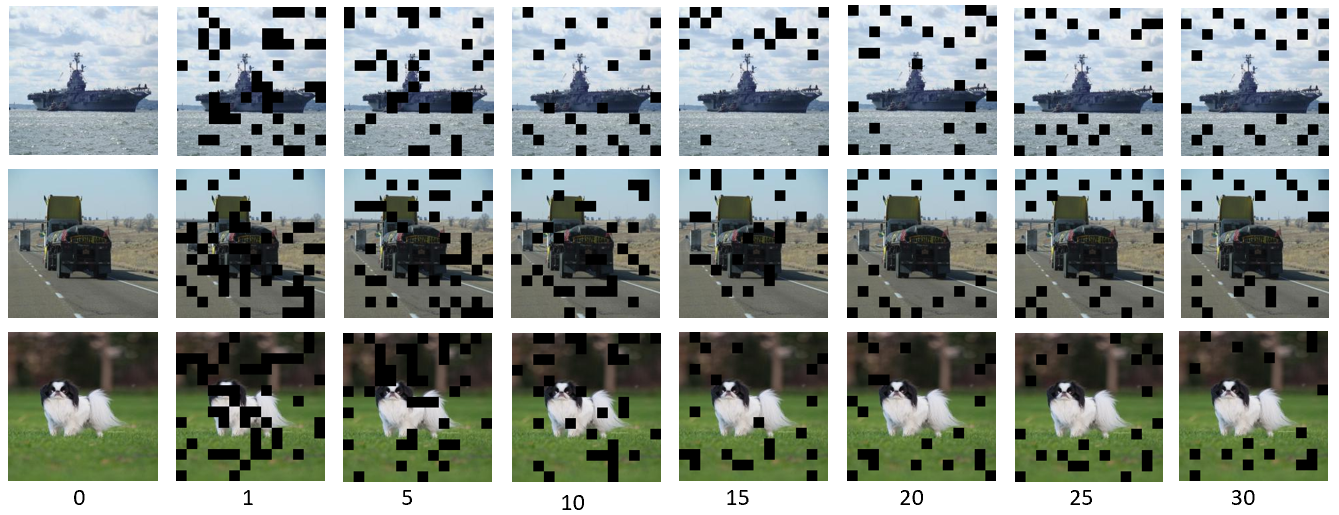}
	\caption{Illustrations of the benign images and their corresponding evolution process of masks. The number under the picture indicates the DE iterations $T_{DE}$. We see that the dropped patches gradually gather in the background while retaining the foreground object.}
    \label{fig:3}
\end{figure*}

\begin{algorithm}[t]
	\renewcommand{\algorithmicrequire}{\textbf{Input:}}
	\renewcommand{\algorithmicensure}{\textbf{Output:}}
	\caption{Attacks Based on Learnable Patch-wise Mask}
	\label{alg:1}
	\begin{algorithmic}[1]
	
		\REQUIRE {Source model $f(\cdot)$, simulated model $s(\cdot)$, clean image $x$ and ground-truth label $y$. population size $P$,  DE iterations $T_{DE}$, patch size $p_s$, hyper-parameter $r$, $\rho$, attack iterations $T_{m}$ for searching population, final attack iterations  $T$.}
		\ENSURE The generated adversarial image $X^{adv}$
		\STATE {\small  Initialize patch-wise masks $\mathcal{M}(0)$ randomly with the zeros' rate of $r$ and patch size of $p_s$, and get the superior individual $\mathcal{M}^*(0)$ by calculate the feedback of $\mathcal{M}(0)$ in Eq. \ref{eq:de_ce}.}
		\FOR{$k$ = $1$  to $T_{DE}$} \label{alg:1:step2}
		    \STATE {\small Generate the candidate generation solutions $\mathcal{M}^{'}(k+1)$: \\
		    { \small Get ${M}_{i}^{'}(k+1)$ by Eq. \ref{eq:de2}, {$i \in(0, \rho * P]$;}} 
            \emph{\small ~//~ Crossover.}  \\
		    {\small Get ${M}_{i}^{'}(k+1)$ by Eq. \ref{eq:de3}, {$i \in(\rho * P, P]$;}} 
                  \emph{\small~ //~ Mutation.} \\
            % {\small combine ${M}_{i}^{'}(k+1)$ and ${M}_{j}^{'}(k+1)$ according to Eq. \ref{eq:de4}}
		    }

            \FOR{$i$ = $1$  to $P$} \label{alg:1:step4} 
    		    \STATE {\small Drop the image $x$ according to mask $M_{i}^{'}(k+1)$;}
    		    
    		    \STATE {\small Generate  $x_{i}^{adv}$ by I-FGSM with $T_{m}$ based on Eq. \ref{eq:lfgsm};} \label{alg:1:step6}
    		    
    		    \STATE {\small Calculate feedback $\phi _i$ of $x_i^{adv}$ on simulated model $s(\cdot)$ according to Eq. \ref{eq:de_ce};}
    
            \ENDFOR
            \STATE {\small Get the next generation solutions $\mathcal{M}(k+1)$:  \\
            {Get \small${M}_{i}(k+1)$ by Eq. \ref{eq:de4}, {$i \in(0, P]$;}}\emph{\small //~ Selection.}}

		\ENDFOR
		
		% \STATE {\small Sort $\mathcal{M}(T)$ in ascending order according to $\phi$ on $s(\cdot)$, and get the best mask $M^*$ in $\mathcal{M}(T)$};

            \STATE {\small Get the final adversarial image $X^{adv}$  with $T$  by I-FGSM on source model $f(\cdot)$ with 
            $x$ modified by $\mathcal{M}^*$ from $\mathcal{M}(k+1)$;}		
		\STATE {\small\textbf{return}  $X^{adv}$}
    % \STATE {//~ Notes: I-FGSM can be replaced by other attack method, e.g., DI-FGSM, MI-FGSM, TI-FGSM.}
 \end{algorithmic}
      \emph{\small //~ Notes: I-FGSM can be replaced by other attack methods, e.g., DI-FGSM, MI-FGSM, TI-FGSM, etc.}
\end{algorithm}

\textbf{Crossover Strategy.} The crossover strategy generates the candidate next-generation individuals based on the superior individuals in the $k$-th generation. To determine the superior individuals in the $k$-th generation $\mathcal{M}^*(k)$, we calculate the feedback $\phi_i$ of each $M_i(k)$, and take those with the best $\rho * P$ as superior individuals. Then, we perform crossover among superior individuals to generate part of the candidate solutions for the $k+1$-th generation. The crossover operation is formalized as:
\begin{equation}\label{eq:de2}
{M}_{i}^{'}(k+1)=I[{M}_{\pi_{1}}^*(k)+{M}_{\pi_{2}}^*(k)],i=1,...\rho * P,
\end{equation}
where ${M}_{i}^{'}(k+1)$ is the $i$-th candidate individual mask solution in the ${k+1}$-th generation.  $M_{\pi_{1}}^*(k)$ and $M_{\pi_2}^*(k)$ are randomly selected from the superior individuals, and $\pi_{1},\pi_{2} \in(0, \rho * P]$.  $\rho$ is the rate of crossover individuals to the total individuals. $I(\cdot)$ is the indicator function, $I(x)=1$ if $x = 2$; $I(x)=0$ if $x = 0$; $I(x) = 0$ or $1$ (with the probability of $0.5$) if  $x = 1$. The crossover strategy helps to find the location of commonality among masks.

\textbf{Mutation Strategy.} The mutation strategy aims to avoid the suboptimal solution and generate another part of the candidate individuals, which can be expressed as:
\begin{equation}\label{eq:de3}
{M}_{i}^{'}(k+1)=E(M_{i}(k), p_m), i=\rho*P,..., P,
\end{equation}
where  $E(\cdot)$ is the mutation function, which randomly changes the value of each patch position in the mask with probability $p_m$. We set $p_m$ to 1, which is equivalent to randomly generating mutation individuals who have no direct connection with the previous individuals. Meanwhile, we keep the number of 0 or 1 in the individuals constant during the mutation strategy to ensure that the drop rate does not change.
 
\textbf{Selection Strategy.}
After the crossover and mutation operation, we select the best $P$ nonrepeating individuals as final $k+1$ generation solutions based on the feedback $\phi$ from the $k$ generation individuals $\mathcal{M}(k)$ and the $k+1$ generation candidate individuals from $\mathcal{M^{'}}(k+1)$, which can be formalized as follows:
\begin{equation}\label{eq:de4}
\mathcal{M}(k+1) = Top(\mathcal{M}(k), \mathcal{M}^{'}(k+1);P),
\end{equation}
where $Top(\cdot;P)$ denotes the selection function and obtains the best $P$ unique individuals based on feedback $\phi$. we repeat the above process for iteration $T_{DE}$. Then we select the final individual $\mathcal{M}(T_{DE})$ as the final solution $\mathcal{M}^*$. 

\begin{figure}[t]
	\centering
	\includegraphics[width=0.95\linewidth]{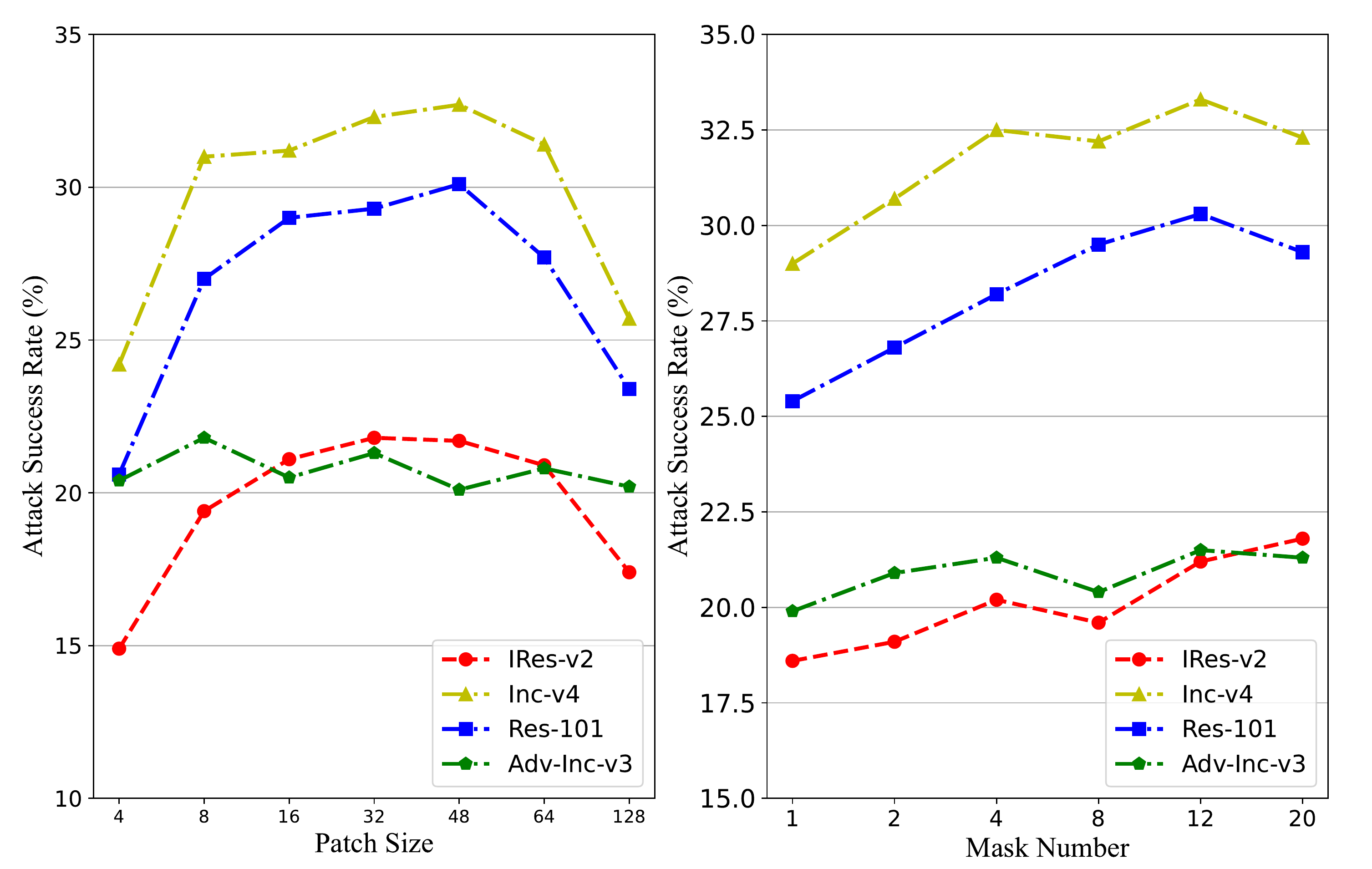}
	\caption{The attack success rates against four models when changing the patch size (left) and the mask number (right). The adversarial images are generated by LPM-I-FGSM on Inc-v3.}
    \label{fig:4}
\end{figure}

\subsection{The Explanation of Model-specific Regions}

We visualize the evolution process of masks and analyze the proposed method. As Figure \ref{fig:3} shows, the dropped positions tend to be stable with continuous evolution. We find that the foreground and its related regions are basically preserved.

Here, we provide a reasonable explanation of this phenomenon. Since the foreground contains the most direct semantic information connection with the label, models will be more inclined to pay attention to the foreground with the label guidance in the training process, leading to the model-generic discriminative regions; In contrast, as a part of the image, the background is utilized in the training process, but this type of information lacks semantic guidance, so different models may focus on different parts of the background information, leading to the model-specific discriminative regions.

This show that the learned mask is meaningful and can drop the model-specific regions, which reduces overfitting and enhances the transferability of the perturbations.  

% add comment

% \subsection{Implementation details}

% In our method, the patch size is a hyperparameter and we explore its influence in the experimental section. Furthermore, to remove the randomness of the predicted mask solution, we predict multiple masks for one image and employ the intersection of multiple learned masks for every step of the final attack. We find that this method can further improve the transfer-based attack success rate. 

\section{Experiments}
% It should be mentioned that all the models used in our paper (including simluated models, source models, and target models) are implemented via the open source manner, which are from pytorch's official pre-training models or the open source model library  timm \cite{rw2019timm}, and our code is based on PyTorch framework.  Due to the different implementations, there may exist small differences with the compared transfer-based attacks versus the  attack success rates in their original papers, however, we ensure that all the performance claimed in the paper are obtained by fair test under the same models, and the entire code and used models will be released after the review process.

\subsection{Experimental Settings}
\textbf{Dataset.} We apply the ImageNet-compatible dataset\footnote{\url{https://github.com/cleverhans-lab/cleverhans/tree/master/cleverhans_v3.1.0/examples/nips17_adversarial_competition/dataset}} following previous work \cite{dong2018boosting,zhang2022improving,xie2019improving}, which contains 1000 images from ImageNet \cite{krizhevsky2012imagenet} with size 299 $\times$ 299 $\times$ 3.

\noindent
\textbf{Models.} The proposed method is evaluated on six normally trained models, including Inception-v3 (Inc-v3) \cite{szegedy2016rethinking}, Inception-v4 (Inc-v4) \cite{szegedy2017inception}, Inception-Resnet-v2 (IRes-v2) \cite{szegedy2017inception}, ResNet-101 (Res-101) \cite{he2016deep} and ResNet-152 (Res-152), and three adversarially trained models, i.e.,  adv-Inception-v3 (adv-Inc-v3), ens4-adv-Inception-v3 (Inc-v3$_{ens4}$) and ens-adv-Inception-ResNet-v2 (IRes-v2$_{ens}$) \cite{DBLP:conf/iclr/TramerKPGBM18}. 

It should be mentioned that all the models used in our paper are implemented in an open source manner, coming from PyTorch's official pretrained models or the open source model library timm \footnote{\url{https://github.com/rwightman/pytorch-image-models}}, and our code is based on the PyTorch framework.  Due to the different implementations, there may be small differences between the compared transfer-based attacks versus the attack success rates in their original papers; however, we ensure that all the performances claimed in the paper are obtained by fair testing under the same models.

In addition, we also evaluate our methods on four other advanced defense strategies, including R\&P \cite{xie2017mitigating}, FD \cite{Liu_2019_CVPR}, RS \cite{jia2019certified},  and NRP \cite{naseer2020self}. The purified images of FD, RS, and  NRP are fed to adv-Inc-v3 to give the final prediction.

\noindent
\textbf{SOTA Methods.} We choose some existing state-of-the-art attacks, including baseline methods I-FGSM \cite{DBLP:journals/corr/KurakinGB16}, advanced gradient calculation methods, i.e., MI-FGSM \cite{dong2018boosting}, NI-FGSM \cite{lin2019nesterov}, and VNI-FGSM \cite{Wang_2021_CVPR}, and input transformation attacks, i.e., TI-FGSM \cite{dong2019evading}, DI-FGSM \cite{xie2019improving}, and S$^2$I-FGSM \cite{long2022frequency}.

\noindent
\textbf{Hyperparameters.} We follow the setting in \cite{dong2018boosting,Wang_2021_CVPR,xie2019improving} with the maximum perturbation $\epsilon$ = 16, the attack iteration $T = 10$, and the step size $\alpha = 1.6/255$ among all experiments. For MI-FGSM, we set $\mu = 1.0$ as recommended in \cite{dong2018boosting}. The diverse probability of DI-FGSM is set as $0.5$. We adopt Gaussian kernel size of $7 \times 7$ for the TI-FGSM. The spectrum transformation number of S$^2$I-FGSM is 20 following the original setting. For our LPM, we set $T_{DE} = 10$, $P = 40$, $\rho = 0.3$, and $r = 0.1$. The number of masks is set to $12$, and the patch size is set to $32$.  The attack iterations $T_{m}$  in searching  model-specific regions is set to 10. Attack Success Rates is used as the evaluation metric, and it refers to the percentage of images that are misclassified by the target model.

\begin{table*}[t] \scriptsize
\caption{The attack success rates (\%) of I-FGSM \cite{DBLP:journals/corr/KurakinGB16}, TI-FGSM \cite{dong2019evading}, DI-FGSM \cite{xie2019improving}, MI-FGSM \cite{dong2018boosting}, S$^2$I-FGSM \cite{long2022frequency}, and enhanced version by our method (LPM). The adversarial examples are generated by \textbf{Inc-v3}. $*$ indicates the white-box model being attacked.}
\centering  

\scalebox{1.02}  { 
\begin{tabular}{
>{\columncolor[HTML]{FFFFFF}}c |
>{\columncolor[HTML]{FFFFFF}}c  :
>{\columncolor[HTML]{FFFFFF}}c 
>{\columncolor[HTML]{FFFFFF}}c 
>{\columncolor[HTML]{FFFFFF}}c 
>{\columncolor[HTML]{FFFFFF}}c 
>{\columncolor[HTML]{FFFFFF}}c 
>{\columncolor[HTML]{FFFFFF}}c 
>{\columncolor[HTML]{FFFFFF}}c 
>{\columncolor[HTML]{FFFFFF}}c 
>{\columncolor[HTML]{FFFFFF}}c 
>{\columncolor[HTML]{FFFFFF}}c 
>{\columncolor[HTML]{FFFFFF}}c 
>{\columncolor[HTML]{EFEFEF}}c }
\hline
Attack       & Inc-v3       & IRes-v2       & Inc-v4        & Res-101       & Res-152       & adv-Inc-v3    & Inc-v3$_{ens4}$ & IRes-v2$_{ens}$ & R\&P          & FD            & RS            & NRP           & Avg            \\ \hline
I-FGSM       & 100$^*$          & 16.7          & 26.7          & 24.1          & 21.5          & 20.3          & 10.6            & \textbf{8.9}             & \textbf{5.9}           & 22.9          & 17.1          & 27.8          & 25.21          \\
\textbf{LPM-I-FGSM}      & \textbf{100}$^*$ & \textbf{21.2} & \textbf{33.3} & \textbf{30.2} & \textbf{28.8} & \textbf{21.5} & \textbf{11.0}     & 8.3             & 5.6           & \textbf{25.2} & \textbf{18.1} & \textbf{28.7} & \textbf{27.66} \\ \hline
TI-FGSM      & 100$^*$          & 16.8          & 24.8          & 21.0            & 19.6          & 20.8          & 11.6            & 9.9             & 7.8           & 26.5          & 18.0            & 27.6          & 25.37          \\
\textbf{LPM-TI-FGSM}     & \textbf{100}$^*$ & \textbf{22.4} & \textbf{34.6} & \textbf{28.6} & \textbf{24.3} & \textbf{22.3} & \textbf{13.8}   & \textbf{11.4}   & \textbf{8.0}    & \textbf{30.7} & \textbf{19.4} & \textbf{28.0}   & \textbf{28.63} \\ \hline
DI-FGSM      & 99.7$^*$     & 29.8          & 46.4          & 33.2          & 32.9          & 22.3          & 12.7            & 10.9            & 7.3           & 27.7          & 17.9          & 28.7          & 30.79          \\
\textbf{LPM-DI-FGSM}     & \textbf{100}$^*$ & \textbf{44.6} & \textbf{59.5} & \textbf{47.7} & \textbf{45.3} & \textbf{25.6} & \textbf{17.4}   & \textbf{12.0}     & \textbf{10.6} & \textbf{32.4} & \textbf{19.6} & \textbf{28.8} & \textbf{36.96} \\ \hline
MI-FGSM      & 100$^*$       & 40.1          & 48.3          & 43.1          & 41.0            & 31.4          & 16.8            & 14.1            & 10.5          & 36.3          & 24.6          & 32.6          & 36.57          \\
\textbf{LPM-MI-FGSM}     & \textbf{100}$^*$ & \textbf{51.5} & \textbf{60.7} & \textbf{56.1} & \textbf{53.7} & \textbf{36.6} & \textbf{20.9}   & \textbf{16.5}   & \textbf{12.4} & \textbf{43.3} & \textbf{26.8} & \textbf{34.4} & \textbf{42.74} \\ \hline
S$^2$I-FGSM  & 99.5$^*$        & 53.1          & 61.6          & 42.0            & 44.3          & 35.3          & 27.2            & 18.5            & 15.9          & 48.5          & 23.9          & 35.3          & 42.09          \\
\textbf{LPM-S$^2$I-FGSM} & \textbf{100}$^*$ & \textbf{67.2} & \textbf{73.0}   & \textbf{58.7} & \textbf{56.9} & \textbf{40.7} & \textbf{32.2}   & \textbf{23.0}     & \textbf{20.8} & \textbf{55.3} & \textbf{25.7} & \textbf{36.1} & \textbf{49.13} \\ \hline
\end{tabular}
} \label{tab:1}
\end{table*}

\begin{table*}[t] \scriptsize
\caption{The attack success rates (\%) of I-FGSM \cite{DBLP:journals/corr/KurakinGB16}, TI-FGSM \cite{dong2019evading}, DI-FGSM \cite{xie2019improving}, MI-FGSM \cite{dong2018boosting}, S$^2$I-FGSM \cite{long2022frequency}, and enhanced version by our method (LPM). The adversarial examples are generated by \textbf{IRes-v2}. $*$ indicates the white-box model being attacked.}
\centering

\scalebox{1.02}  { 

% \begin{table}[]
\begin{tabular}{
>{\columncolor[HTML]{FFFFFF}}c |
>{\columncolor[HTML]{FFFFFF}}c  :
>{\columncolor[HTML]{FFFFFF}}c 
>{\columncolor[HTML]{FFFFFF}}c 
>{\columncolor[HTML]{FFFFFF}}c 
>{\columncolor[HTML]{FFFFFF}}c 
>{\columncolor[HTML]{FFFFFF}}c 
>{\columncolor[HTML]{FFFFFF}}c 
>{\columncolor[HTML]{FFFFFF}}c 
>{\columncolor[HTML]{FFFFFF}}c 
>{\columncolor[HTML]{FFFFFF}}c 
>{\columncolor[HTML]{FFFFFF}}c 
>{\columncolor[HTML]{FFFFFF}}c 
>{\columncolor[HTML]{EFEFEF}}c }
\hline
Attack   & IRes-v2    & Inc-v3            & Inc-v4          & Res-101         & Res-152         & adv-Inc-v3      & Inc-v3$_{ens4}$ & IRes-v2$_{ens}$ & R\&P            & FD              & RS              & NRP             & Avg             \\ \hline
I-FGSM    & 99.7$^*$    & 24.3              & 23.8            & 19.8            & 17.5            & 21.7            & 9.8             & \textbf{10.7}            & 5.5             & 23.9            & 17.0              & 27.5            & 25.10            \\
\textbf{LPM-I-FGSM}   & \textbf{100$^*$}     & \textbf{27.2}    & \textbf{26.5}   & \textbf{23.0}     & \textbf{20.0}     & \textbf{21.8}   & \textbf{11.1}   & 10.6   & \textbf{6.7}    & \textbf{26.5}   & \textbf{18.4}   & \textbf{28.7}   & \textbf{26.71}  \\ \hline
TI-FGSM    & 99.5$^*$  & 26.6                 & 26.3            & 19.4            & 18.4            & 22.6            & 11.2            & 13.0              & 7.6             & 30.4            & 17.6            & 27.9            & 26.71           \\
\textbf{LPM-TI-FGSM}   & \textbf{100$^*$}   & \textbf{30.5}   & \textbf{30.3}   & \textbf{24.5}   & \textbf{21.4}   & \textbf{23.7}   & \textbf{12.3}   & \textbf{14.0}     & \textbf{8.0}      & \textbf{32.5}   & \textbf{19.4}   & \textbf{29.4}   & \textbf{28.83}  \\ \hline
DI-FGSM   & 98.4$^*$    & 39.7                 & 41.5            & 28.6            & 27.2            & 25.0              & 13.9            & 13.5            & 8.1             & 32.0              & 18.7            & 29.6            & 31.35           \\
\textbf{LPM-DI-FGSM}   & \textbf{100$^*$}   & \textbf{49.7}    & \textbf{52.4}   & \textbf{37.2}   & \textbf{35.4}   & \textbf{27.1}   & \textbf{15.8}   & \textbf{16.1}   & \textbf{10.7}   & \textbf{36.3}   & \textbf{20.5}   & \textbf{30.0}     & \textbf{35.93}  \\ \hline
MI-FGSM     & 99.6$^*$    & 49.8                & 48.0              & 44.5            & 40.2            & 34.5            & 18.3            & 20.6            & 11.6            & 39.6            & 27.5            & 34.4            & 39.05           \\
\textbf{LPM-MI-FGSM}   & \textbf{100$^*$}   & \textbf{63.1}   & \textbf{59.2}   & \textbf{54.7}   & \textbf{50.9}   & \textbf{38.2}   & \textbf{21.4}   & \textbf{23.8}   & \textbf{14.2}   & \textbf{45.8}   & \textbf{29.6}   & \textbf{34.9}   & \textbf{44.65}  \\ \hline
S$^2$I-FGSM     & 99.2$^*$     & 59.2        & 58.2            & 41.0              & 40.6            & 40.2            & \textbf{27.7}            & 28.3            & 18.6            & 54.3            & 25.0              & 36.3            & 44.05           \\
\textbf{LPM-S$^2$I-FGSM }& \textbf{99.9$^*$}   & \textbf{64.7}   & \textbf{64.1}   & \textbf{49.3}   & \textbf{48.7}   & \textbf{43.8}   & 27.4   & \textbf{29.6}   & \textbf{21.6}   & \textbf{58.6}   & \textbf{26.8}   & \textbf{36.5}   & \textbf{47.58}  \\ \hline
\end{tabular}
} \label{tab:2}
\end{table*}

\begin{table*}[t] \scriptsize
\caption{The attack success rates (\%) of I-FGSM \cite{DBLP:journals/corr/KurakinGB16}, TI-FGSM \cite{dong2019evading}, DI-FGSM \cite{xie2019improving}, MI-FGSM \cite{dong2018boosting}, S$^2$I-FGSM \cite{long2022frequency}, and enhanced version by our method (LPM). The adversarial examples are generated by \textbf{Res-152}. $*$ indicates the white-box model being attacked.}
\centering  

\scalebox{1.02}  { 

% \begin{table}[]
\begin{tabular}{
>{\columncolor[HTML]{FFFFFF}}c |
>{\columncolor[HTML]{FFFFFF}}c  :
>{\columncolor[HTML]{FFFFFF}}c 
>{\columncolor[HTML]{FFFFFF}}c 
>{\columncolor[HTML]{FFFFFF}}c 
>{\columncolor[HTML]{FFFFFF}}c 
>{\columncolor[HTML]{FFFFFF}}c 
>{\columncolor[HTML]{FFFFFF}}c 
>{\columncolor[HTML]{FFFFFF}}c 
>{\columncolor[HTML]{FFFFFF}}c 
>{\columncolor[HTML]{FFFFFF}}c 
>{\columncolor[HTML]{FFFFFF}}c 
>{\columncolor[HTML]{FFFFFF}}c 
>{\columncolor[HTML]{EFEFEF}}c }
\hline
Attack     & Res-152   & Inc-v3          & IRes-v2         & Inc-v4          & Res-101                   & adv-Inc-v3      & Inc-v3$_{ens4}$ & IRes-v2$_{ens}$ & R\&P            & FD              & RS              & NRP             & Avg             \\ \hline
I-FGSM     & 100$^*$   & 27.8            & 14.6            & 26.1            & 87.5                     & 19.9            & 10.8            & 8.8             & 7.0               & 22.2            & 19.0              & 27.8            & 30.96           \\
\textbf{LPM-I-FGSM}  & \textbf{100$^*$}    & \textbf{34.0}     & \textbf{18.3}   & \textbf{31.5}   & \textbf{95.7}    & \textbf{20.7}   & \textbf{11.2}   & \textbf{9.8}    & \textbf{7.3}    & \textbf{23.0}     & \textbf{19.9}   & \textbf{28.5}   & \textbf{33.33}  \\ \hline
TI-FGSM   & 100$^*$     & 31.5            & 17.5            & 26.7            & 88.2                    & 22.2            & 12.4            & 10.4            & 10.0              & 28.4            & 22.3            & 28.0              & 33.13           \\
\textbf{LPM-TI-FGSM}   & \textbf{100$^*$}   & \textbf{41.0}     & \textbf{24.8}   & \textbf{37.7}   & \textbf{95.6}   & \textbf{23.9}   & \textbf{16.2}   & \textbf{12.8}   & \textbf{11.3}   & \textbf{31.2}   & \textbf{24.6}   & \textbf{29.8}   & \textbf{37.41}  \\ \hline
DI-FGSM    & 100$^*$     & 53.3            & 33.3            & 51.9            & 95.8                    & 23.6            & 14.9            & 11.8            & 11.4            & 28.6            & 21.9            & 29.4            & 39.66           \\
\textbf{LPM-DI-FGSM}     & \textbf{100$^*$}    & \textbf{71.5}   & \textbf{45.9}   & \textbf{69.3}   & \textbf{99.5}   & \textbf{25.9}   & \textbf{18.8}   & \textbf{14.7}   & \textbf{14.1}   & \textbf{32.0}     & \textbf{24.1}   & \textbf{30.7}   & \textbf{45.54}  \\ \hline
MI-FGSM     & 100$^*$    & 52.2            & 36.7            & 46.8            & 96.3                    & 30.2            & 18.3            & 15.0              & 12.4            & 32.8            & 31.1            & 34.1            & 42.16           \\
\textbf{LPM-MI-FGSM}     & \textbf{100$^*$}    & \textbf{64.9}   & \textbf{45.7}   & \textbf{57.8}   & \textbf{98.9}  & \textbf{31.6}   & \textbf{20.7}   & \textbf{17.6}   & \textbf{14.9}   & \textbf{37.4}   & \textbf{34.1}   & \textbf{36.6}   & \textbf{46.68}  \\ \hline
S$^2$I-FGSM   & 100$^*$  & 72.6            & 62.8            & 71.4            & 98.7                   & 41.3            & 33.8            & 27.7            & 29.5            & 53.3            & 33.7            & 37.5            & 55.19           \\
\textbf{LPM-S$^2$I-FGSM}   & \textbf{100$^*$}   & \textbf{78.2}            & \textbf{65.9}            & \textbf{76.3}            & \textbf{99.2}                & \textbf{43.4}            & \textbf{36.1}            & \textbf{32.6}            & \textbf{31.6}            & \textbf{56.7}            & \textbf{34.6}            & \textbf{38.6}            & \textbf{57.77}    \\ \hline
\end{tabular}
} \label{tab:3}
\end{table*}

\begin{table*}[t] \scriptsize
% \scriptsize
\caption{The attack success rates (\%) toward seven defense methods in the ensemble attack setting. DTS denotes the combination of TI-FGSM \cite{dong2019evading}, DI-FGSM \cite{xie2019improving}, and SI-FGSM \cite{lin2019nesterov}. The adversarial examples are generated via an ensemble of \textbf{Inc-v3}, \textbf{IRes-v2}, and \textbf{Res-101},  the weight of each is 1/3. }
% \caption{The attack success rates (\%) of MI-FGSM-DTS, NI-FGSM-DTS, VMI-FGSM-DTS, and LMI-FGSM-DTS under a single-model setting, $*$ indicates the white-box model being attacked. The best results are marked in bold.}
% \vspace{-0.25cm}
\centering  

\scalebox{1.15}
{

\begin{tabular}{
>{\columncolor[HTML]{FFFFFF}}c |
>{\columncolor[HTML]{FFFFFF}}c 
>{\columncolor[HTML]{FFFFFF}}c 
>{\columncolor[HTML]{FFFFFF}}c 
>{\columncolor[HTML]{FFFFFF}}c 
>{\columncolor[HTML]{FFFFFF}}c 
>{\columncolor[HTML]{FFFFFF}}c 
>{\columncolor[HTML]{FFFFFF}}c 
>{\columncolor[HTML]{FFFFFF}}c 
>{\columncolor[HTML]{FFFFFF}}c 
>{\columncolor[HTML]{EFEFEF}}c }
\hline
        Attack            & Inc-v4        & Res-152       & adv-Inc-v3    & Inc-v3$_{ens4}$ & IRes-v2$_{ens}$ & R\&P          & FD            & RS            & NRP           & Avg            \\ \hline
MI-FGSM-DTS         & 94.5          & 98.0            & 83.4          & 87.9            & 80.5            & 85.6          & 93.1          & 78.5          & 55.0            & 80.57          \\
\textbf{LPM-MI-FGSM-DTS }    & \textbf{98.1} & \textbf{99.5} & \textbf{89.8} & \textbf{93.1}   & \textbf{87.0}     & \textbf{90.5} & \textbf{96.7} & \textbf{86.1} & \textbf{58.5} & \textbf{85.96} \\ \hline
NI-FGSM-DTS         & 96.2          & 98.5          & 83.7          & 90.1            & 82.1            & 85.3          & 93.5          & 78.7          & 41.8          & 79.31          \\
\textbf{LPM-NI-FGSM-DTS}     & \textbf{98.2} & \textbf{99.7} & \textbf{88.6} & \textbf{94.3}   & \textbf{87.7}   & \textbf{91.7} & \textbf{96.2} & \textbf{85.9} & \textbf{57.5} & \textbf{85.99} \\ \hline
VNI-FGSM-DTS        & 93.9          & 96.7          & 89.6          & 90.9            & 85.2            & 86.9          & 96.3          & \textbf{86.5} & 74.5          & 87.13          \\
\textbf{LPM-VNI-FGSM-DTS}    & \textbf{95.9} & \textbf{98.0}   & \textbf{92.3} & \textbf{92.6}   & \textbf{88.1}   & \textbf{88.2} & \textbf{96.7} & 85.1          & \textbf{78.0}   & \textbf{88.71} \\ \hline
S$^2$I-MI-FGSM-DTS     & 96.9          & 97.5          & 95.8          & 95.6            & 92.7            & 94.1          & 97.7          & 92.4          & \textbf{79.9} & 92.60           \\
\textbf{LPM-S$^2$I-MI-FGSM-DTS} & \textbf{98.0}   & \textbf{97.8} & \textbf{96.3} & \textbf{96.4}   & \textbf{93.1}   & \textbf{94.5} & \textbf{98.2} & \textbf{94.3} & 78.3          & \textbf{93.01} \\ \hline
\end{tabular}

} \label{tab:4}
\end{table*}

\begin{figure}
	\centering
    % \vspace{-0.8em}
	\includegraphics[width=\linewidth]{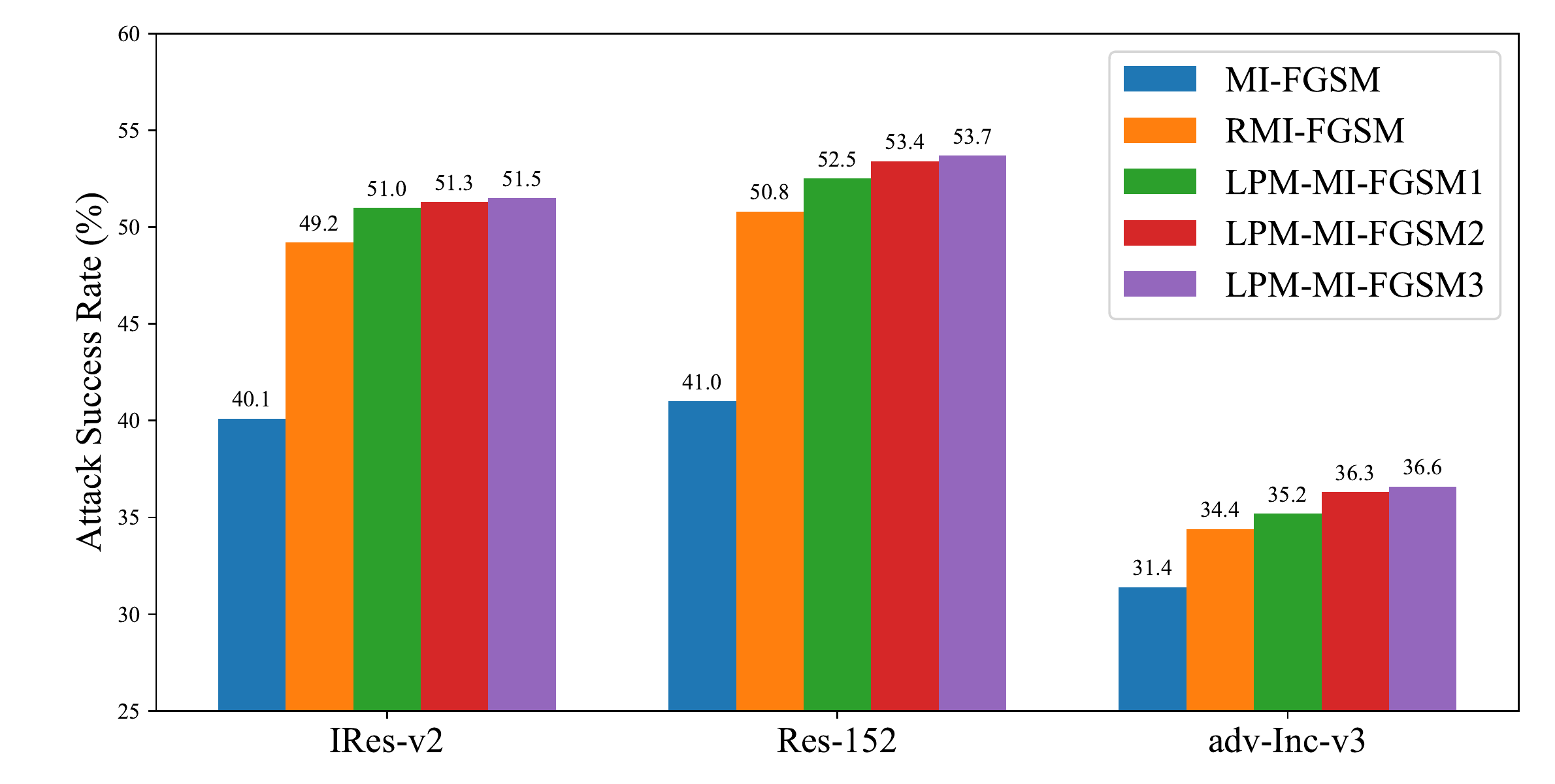}
	\caption{The attack success rates against four models with adversarial images generated on Inc-v3. RI-FGSM represents I-FGSM with random patch-wise masks. LPM-I-FGSM1, LPM-I-FGSM2, LPM-I-FGSM3 represent I-FGSM with learned patch-wise masks, and the numbers of simulated models are 1, 2, and 3, respectively.}
    \label{fig:5}
    % \vspace{-0.8em}
\end{figure}

% \subsection{Ablation Study}
% \subsection{The Influence of Patch Size}
\subsection{Effect of Hyperparameters in the Proposed Method}

In the experiments, we find that the patch size plays an important role in boosting transferability, and we explore its effect on the results. The curves of different patch sizes are shown in Figure \ref{fig:4} (left). The patch size of $1$, i.e., pixel-level drop, has a negative effect on transferability. We think that pixel-level drop is not effective enough to prune the model-specific regions, resulting in poor gradients. As the patch size increases within a certain range, the attack success rate is continuously improved.  Based on the results, the patch size is set to 32 in later experiments.

In addition to the patch size $p_s$, the number of masks is also an important hyperparameter. Figure \ref{fig:4} (right) shows the results of adjusting the mask number (i.e., how many patch-wise masks to learn for each image by using the differential evolution algorithm). It can be found that the greater the mask number is, the stronger the transferability of generated adversarial examples. Considering the computational complexity and attack success rate, we set the number of masks to 12.

\subsection{Ablation Study}

To explore the effectiveness of the proposed learning strategy based on the differential evolution algorithm, we conducted ablation studies. The experimental results are shown in Figure \ref{fig:5}. RMI-FGSM represents MI-FGSM with random patch-wise masks (randomly generate a set of masks with the same number and size of LPM, other settings are the same as LPM-MI-FGSM as described in Section IV.A.). LPM-MI-FGSM1, LPM-MI-FGSM2, and LPM-MI-FGSM3 represent MI-FGSM with learnable patch-wise masks, and the numbers of simulated models are 1 (ResNet-50), 2 (ResNet-50 and VGG-16), and 3 (ResNet-50, VGG-16, and DenseNet-161), respectively. As seen from the results, attacks with learned patch-wise masks outperform random masks and no masks, which demonstrates the effectiveness of the proposed learnable strategy for searching model-specific regions on images and improving adversarial transferability. For example, the success rate of RMI-FGSM is 34.4\% on adv-Inc-v3, while that of the proposed learning-based attack LPM-MI-FGSM1 is 35.2\%. As the number of simulated models increases, the attack success rate of  LPM-MI-FGSM1 is 36.6\% on adv-Inc-v3. The results for the IRes-v2 and Res-152 also follow this trend.

\subsection{Comparison of Transferability}
We conduct experiments based on the existing advanced methods (i.e.,I-FGSM \cite{DBLP:journals/corr/KurakinGB16}, TI-FGSM \cite{dong2019evading}, DI-FGSM \cite{xie2019improving}, MI-FGSM \cite{dong2018boosting}, and S$^2$I-FGSM \cite{long2022frequency}) and enhance these methods with our LPM. The results are reported in Table \ref{tab:1}, Table \ref{tab:2}, and Table \ref{tab:3}. From the table, we can see that the LPM can effectively strengthen the existing SOTA methods when attacking white-box models. More importantly, the proposed method can significantly improve the success rate of transfer attacks. Specifically, when we generate adversarial examples using Inc-v3 as the white-box model, LPM-I-FGSM, LPM-TI-FGSM, LPM-DI-FGSM, LPM-MI-FGSM, and LPM-S$^2$I-FGSM achieve average transfer success rates of 27.66\%, 28.63\%, 36.96\%, 42.74\%, and 49.13\%, which outperform the original attack performance of 2.45\%, 3.26\%, 6.17\%, 6.17\%, and 7.04\% respectively. This reveals that our LPM can further improve the adversarial transferability of existing methods.

\subsection{Combined with Existing Methods}

\begin{figure}
	\centering
    % \vspace{-0.8em}
	\includegraphics[width=0.9\linewidth]{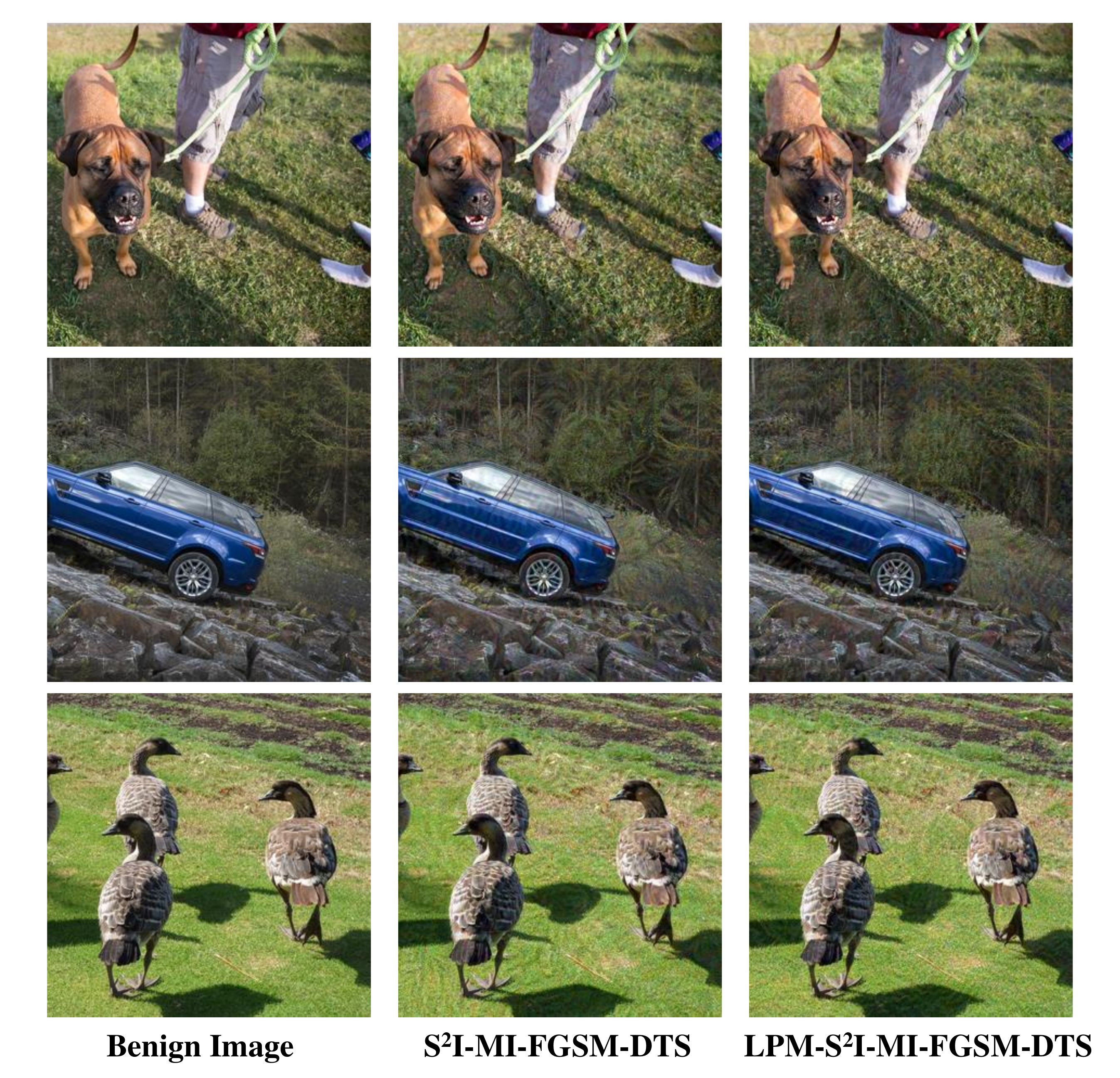}
	\caption{Benign images (left) and the adversarial examples generated by S$^2$I-MI-FGSM-DTS (middle) and our LPM-S$^2$I-MI-FGSM-DTS (right).}
    \label{fig:6}
    % \vspace{-0.8em}
\end{figure}

The momentum term can stabilize update directions and improve adversarial transferability \cite{dong2018boosting}. Lin et al. \cite{lin2019nesterov} show that the combination (DTS) of translation-invariant (TI) \cite{dong2019evading}, diverse inputs (DI) \cite{xie2019improving}, and scale invariant (SI) \cite{lin2019nesterov} can further improve attack success rates. Liu et al. \cite{liu2016delving} find that attacking multiple models simultaneously can effectively improve the transferability of adversarial images. Here, we combine all of them with existing SOTA methods and adopt a model ensemble attack by averaging the logit outputs of Inc-v3, IRes-v2, and ResNet-152 to calculate the loss to generate adversarial examples, and the results are reported in Table \ref{tab:4}.
We can observe that our LPM can further enhance the adversarial transferability for existing methods MI-FGSM-DTS \cite{dong2018boosting}, NI-FGSM-DTS \cite{lin2019nesterov}, VNI-FGSM-DTS \cite{Wang_2021_CVPR}, and S$^2$I-MI-FGSM-DTS \cite{long2022frequency}, which can achieve an increase in the attack success rate of 5.39\%, 6.68\%, 1.58\%, and 0.41\%, respectively. This is a remarkable improvement and indicates that our approach has good scalability and can be combined with existing methods to further improve transferability. The effect of the attack is shown in Figure \ref{fig:6}.

\section{Attack on Vision Transformer Model}

To comprehensively evaluate the proposed method, we conduct experiments on vision transformers (ViTs). Eight ViTs are selected, i.e., ViT-B \cite{dosovitskiy2020image}, PiT-B \cite{heo2021rethinking}, DeiT-B \cite{touvron2021training}, ViT-S \cite{dosovitskiy2020image}, PiT-S \cite{heo2021rethinking}, Visformer-S \cite{chen2021visformer}, LeViT-256 \cite{graham2021levit}, and ConViT-B \cite{d2021convit}. All ViTs are pretrained on ImageNet. We adopt model ensemble attack and use Inc-v3, IRes-v2, and Res-101 as white-box models to generate adversarial examples. The results are reported in Table \ref{tab:sub}. We can see that LPM can effectively improve the average attack success rate for the existing methods MI-FGSM-DTS \cite{dong2018boosting}, NI-FGSM-DTS \cite{lin2019nesterov}, VNI-FGSM-DTS \cite{Wang_2021_CVPR}, and S$^2$I-MI-FGSM-DTS \cite{long2022frequency}, which can achieve an increase in the attack success rate of 7.29\%, 6.13\%,  1.85\%, and 0.55\%, respectively. The experimental results in Table \ref{tab:sub} illustrate the applicability of the proposed method to ViTs, and indicate that the current ViTs are prone to the threat of adversarial examples.

\subsection{Trade-off between Performance and Efficiency}

In this subsection, we discuss the trade-off between the performance and efficiency of the LPM. As a preprocessing method, the time consumption of LPM is mainly concentrated in the process of searching the model-specific regions (DE). The time complexity of the DE algorithm is determined by these hyperparameters: the DE iterations $T_{DE}$ (step \ref{alg:1:step2} in Algorithm \ref{alg:1}), the population size $P$ (step \ref{alg:1:step4} in Algorithm \ref{alg:1}), and the attack iterations $T_{m}$ (step \ref{alg:1:step6} in Algorithm \ref{alg:1}). Because the time cost caused by $P$ can be avoided by parallelizing the processing of the population, we mainly discuss the impact of the DE iterations $T_{DE}$  and the attack iterations $T_{m}$. The $T_{DE}$ and  $T_{m}$  determine whether proper individual masks can be selected and alleviate the instability of the performance caused by randomness. The experiments are conducted on LPM-I-FGSM with NVIDIA GeForce RTX 3080.

\begin{table*}[ht] \scriptsize
% \scriptsize
\caption{The attack success rates (\%) toward ViTs in the ensemble attack setting. DTS denotes the combination of TI \cite{dong2019evading}, DI \cite{xie2019improving}, and SI \cite{lin2019nesterov}. The adversarial examples are generated via an ensemble of \textbf{Inc-v3}, \textbf{IRes-v2}, and \textbf{Res-101},  the weight of each is 1/3. }
\centering  

\scalebox{1.15}
{

% \vspace{-0.25cm}

\begin{tabular}{
>{\columncolor[HTML]{FFFFFF}}c |
>{\columncolor[HTML]{FFFFFF}}c 
>{\columncolor[HTML]{FFFFFF}}c 
>{\columncolor[HTML]{FFFFFF}}c 
>{\columncolor[HTML]{FFFFFF}}c 
>{\columncolor[HTML]{FFFFFF}}c 
>{\columncolor[HTML]{FFFFFF}}c 
>{\columncolor[HTML]{FFFFFF}}c 
>{\columncolor[HTML]{FFFFFF}}c 
>{\columncolor[HTML]{EFEFEF}}c }
\hline
Attack                       & ViT-B         & ViT-S         & DeiT-S        & PiT-B         & PiT-S         & Visformer-S   & LeViT-256     & ConViT-B      & Avg            \\ \hline
MI-FGSM-DTS                  & 67.2          & 79.1          & 64.4          & 75            & 83.2          & 87.8          & 87.2          & 68.2          & 76.51          \\
\textbf{LPM-MI-FGSM-DTS}     & \textbf{74.8} & \textbf{85.2} & \textbf{73.4} & \textbf{82.6} & \textbf{89.7} & \textbf{93.4} & \textbf{93.7} & \textbf{77.6} & \textbf{83.80} \\ \hline
NI-FGSM-DTS                  & 67.3          & 78.8          & 66.0          & 75.6          & 84.4          & 90.0          & 89.3          & 69.6          & 77.63          \\
\textbf{LPM-NI-FGSM-DTS}     & \textbf{74.1} & \textbf{85.0} & \textbf{73.3} & \textbf{82.6} & \textbf{89.8} & \textbf{94.2} & \textbf{94.6} & \textbf{76.5} & \textbf{83.76} \\ \hline
VNI-FGSM-DTS                 & 77.0          & 84.5          & 64.2          & 68.7          & 81.5          & 84.8          & 85.2          & 69.0          & 76.86          \\
\textbf{LPM-VNI-FGSM-DTS}    & \textbf{78.9} & \textbf{86.1} & \textbf{65.4} & \textbf{70.9} & \textbf{83.6} & \textbf{86.6} & \textbf{88.5} & \textbf{69.7} & \textbf{78.71} \\ \hline
S$^2$I-MI-FGSM-DTS              & 83.6          & 89.9          & 76.7          & 83.2          & 89.6          & 92.1          & 92.9          & 79.3          & 85.91          \\
\textbf{LPM-S$^2$I-MI-FGSM-DTS} & \textbf{84.2} & \textbf{90.8} & \textbf{77.4} & \textbf{84.1} & \textbf{89.9} & \textbf{92.5} & \textbf{93.2} & \textbf{79.6} & \textbf{86.46} \\ \hline
\end{tabular}

} \label{tab:sub}
\end{table*}

\begin{table}[th] \scriptsize
% \scriptsize
\caption{The attack success rates (\%) and time cost (s) with different  $T_{DE}$ and  $T_{m}$ of LPM, other setting is the same as mentioned before. The time cost refers to the time applying a single image to search for a learnable mask. The adversarial examples are generated by Inc-v3. $T_{DE} = 0$ denotes that we apply  generated Patch-wise masks while the iteration of DE is 0, which is euqal to random generated masks.}
\centering  
\scalebox{1.07}
{
\begin{tabular}{
>{\columncolor[HTML]{FFFFFF}}c 
>{\columncolor[HTML]{FFFFFF}}c |
>{\columncolor[HTML]{FFFFFF}}c 
>{\columncolor[HTML]{FFFFFF}}c 
>{\columncolor[HTML]{FFFFFF}}c 
>{\columncolor[HTML]{FFFFFF}}c |
>{\columncolor[HTML]{FFFFFF}}c }
\hline
$T_{DE}$  &  $T_{m}$  & IRes-v2 & Inc-v4 & Res-101 & Res-152 &  Time (s) \\ \hline
0  & - & 19.3     &   30.8    &  27.4   &    23.8        &   2.7      \\
5  & 5 &  21.2  & 32.9  &  28.7  &  25.8   & 3.8      \\
\textbf{5}  & \textbf{8} &  21.2  &  35.0  &   29.8   &   27.0 &   6.5      \\
5  & 10 & 22.9    & 32.5    &    28.5  & 26.7     & 8.3    \\
10  & 5 &  22.1  & 33.1 & 29.8 & 25.6  &  7.5     \\
10  & 8 &   22.1 & 34.1 & 30.9 & 27.0 & 11.9     \\
10 & 10 & 21.2    & 33.3    &    30.2  & 28.8      & 14.9    \\ \hline
\end{tabular}
} \label{time}
\vspace{-0.35cm}
\end{table}

The results in Table \ref{time} show that $T_{DE}$ and  $T_{m}$ will obviously affect the time cost. When $T_{DE}$ and $T_{m}$ decrease, the time cost will correspondingly decrease. Meanwhile, the attack performance will have a minor fluctuation in a certain range and still maintain a competitive performance, e.g.,  $T_{DE}=5$ and  $T_{m}=8$ is a cost-effective solution compared with the original setting. Therefore, our LPM can be chosen with different hyperparameter settings in the DE algorithm according to the actual requirement. 
% We do not deny that more appropriate parameters exist to achieve better performances than that claimed in the paper.

\subsection{Quantification of Gradient Aggregation}
To measure the impact of our method on gradient aggregation, we introduce the clustering coefficient \cite{watts1998collective,holland1971transitivity} to measure the aggregation degree of saliency maps \cite{smilkov2017smoothgrad} for clean images and images masked by our DE algorithm.

Figure \ref{fig:1}(a) shows that owing to model-specific regions, the saliency maps show a \textbf{cluttered} distribution. When these model-specific regions are masked, the saliency maps  become \textbf{clustered} (Figure \ref{fig:1}(b)). Therefore, we can use the aggregation degree of saliency maps in an image to reflect whether our hypothesis about the model-specific is true. Here we introduce the local clustering coefficient \cite{watts1998collective,holland1971transitivity} to compute quantitative evidence. In saliency maps, if one pixel is greater than a certain threshold, it is seen as a vertex. We compute $C_i$ for each vertex, which denotes how close one vertex’s neighbors are to being a clique. The $C_i$ can be defined as follows:
\begin{equation}
 C=\frac{1}{N}\sum_{i=1}^{N}{C_i}, ~
C_i = \frac{2|\{e_{jk}: v_j,v_k \in L_i\}|}{k_i(k_i-1)},
\end{equation}
where $N$ is the total vertex, and $L_i$ represents the set of immediately connected vertices with vertex $v_i$. $v_j$ and $v_k$ are two vertices that belong to $L_i$. $e_{jk}$ denotes the edge  connecting vertex $v_j$ with vertex $v_k$, and $k_i$ denotes the number of neighbors of vertex $v_i$. $\vert\cdot\vert$ denotes the number of edges. 

Here we also compute the average clustering coefficient of saliency maps generated by three DNN models, i.e., ResNet-50, Inception-v3, and VGG-16. The result in Table \ref{tab:quantification} shows a higher aggregation degree of mask images than benign images, which means that our masks prune model-specific regions and lead to clustered saliency maps.

\begin{table}[htb]
\caption{The average clustering coefficient of three DNN models.}
\label{tab:quantification}
\centering 
\begin{tabular}{c|c|c|c}
\hline
Saliency map & Inc-v3 & VGG-16 & Res-50  \\ \hline 
Benign image &0.179    &0.153   &0.20           \\ \hline
Masked image  &\textbf{0.212}   &\textbf{0.190}   &\textbf{0.23}           \\ \hline
\end{tabular}
\end{table}

%Fig.1(a) shows that owing to model-specific regions, the saliency maps show a \textbf{cluttered} distribution. When these model-specific regions are masked, the saliency maps  become \textbf{clustered} (Fig.1(b)). Therefore, we can use the aggregation degree of saliency maps in an image to reflect whether our idea works. For that, we introduce local clustering coefficient\footnote{\tiny Duncan J Watts, \emph{et al.} Collective dynamics of ‘small-world’networks. nature, 393:440–442,1998} to compute a quantitative evidence. In saliency maps, if one pixel is white, it is seen as a vertex. We compute $C_i$ for each vertex, which denotes how close one vertex’s neighbours are to being a clique.
%\vspace{-0.35cm}
%\begin{equation}
%\scriptsize
%  C=\frac{1}{N}\sum_{i=1}^{N}{C_i}, \quad\quad C_i = \frac{2|\{e_{jk}: v_j,v_k \in L_i\}|}{k_i(k_i-1)}
%  \vspace{-0.35cm}
%\end{equation}
%where $N$ is the total vertex, $L_i$ represents the set of immediately connected vertices with vertex $v_i$. $v_j$ and $v_k$ are two vertices that belong to $L_i$. $e_{jk}$ denotes the edge  connecting vertex $v_j$ with vertex $v_k$, $k_i$ denotes the number of neighbours of vertex $v_i$. $\vert\cdot\vert$ denotes the number of edges. 

\section{Conclusion}
In this paper, we argued that model-specific discriminative regions caused overfitting to the source model and thus reduced adversarial transferability. Then, we proposed a learning strategy based on the differential evolution algorithm to search for the patch-wise mask (LPM), which was used to prune model-specific regions when calculating adversarial perturbations. LPM as a preprocessing operation could be integrated with existing gradient-based methods and effectively improve these methods' transfer attack success rates. In the ensemble attack setting, the proposed approach achieved an average success rate of 93.01\% against seven advanced defense mechanisms, demonstrating the effectiveness of our method. 

\section*{Acknowledgement}
This work was supported by the Project of the National Natural Science Foundation of China (No.62076018), and the Fundamental Research Funds for the Central Universities.

% \newpage
\vspace{-1cm}
\begin{IEEEbiography}[{\includegraphics[width=1in,height=1.25in,clip,keepaspectratio]{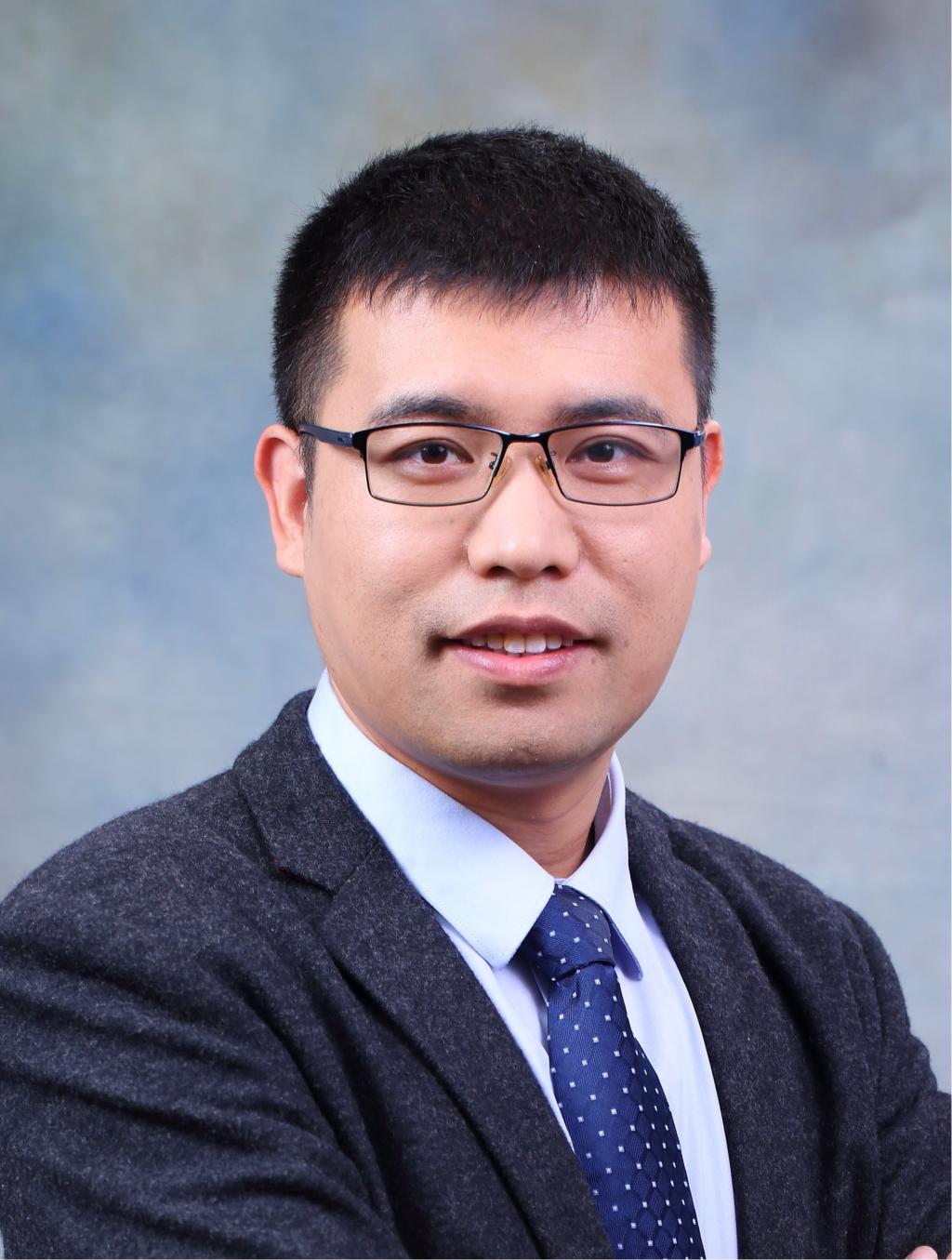}}]{Xingxing Wei} received his Ph.D degree in computer science from Tianjin University, and B.S.degree in Automation from Beihang University (BUAA), China. He is now an Associate Professor at Beihang University (BUAA). His research interests include computer vision, adversarial machine learning and its applications to multimedia content analysis. He is the author of referred
journals and conferences in IEEE TPAMI, TMM, TCYB, TGRS, IJCV, PR, CVIU, CVPR, ICCV, ECCV, ACMMM, AAAI, IJCAI etc.
\end{IEEEbiography}
\vspace{-1cm}
\begin{IEEEbiography}[{\includegraphics[width=1in,height=1.25in,clip,keepaspectratio]{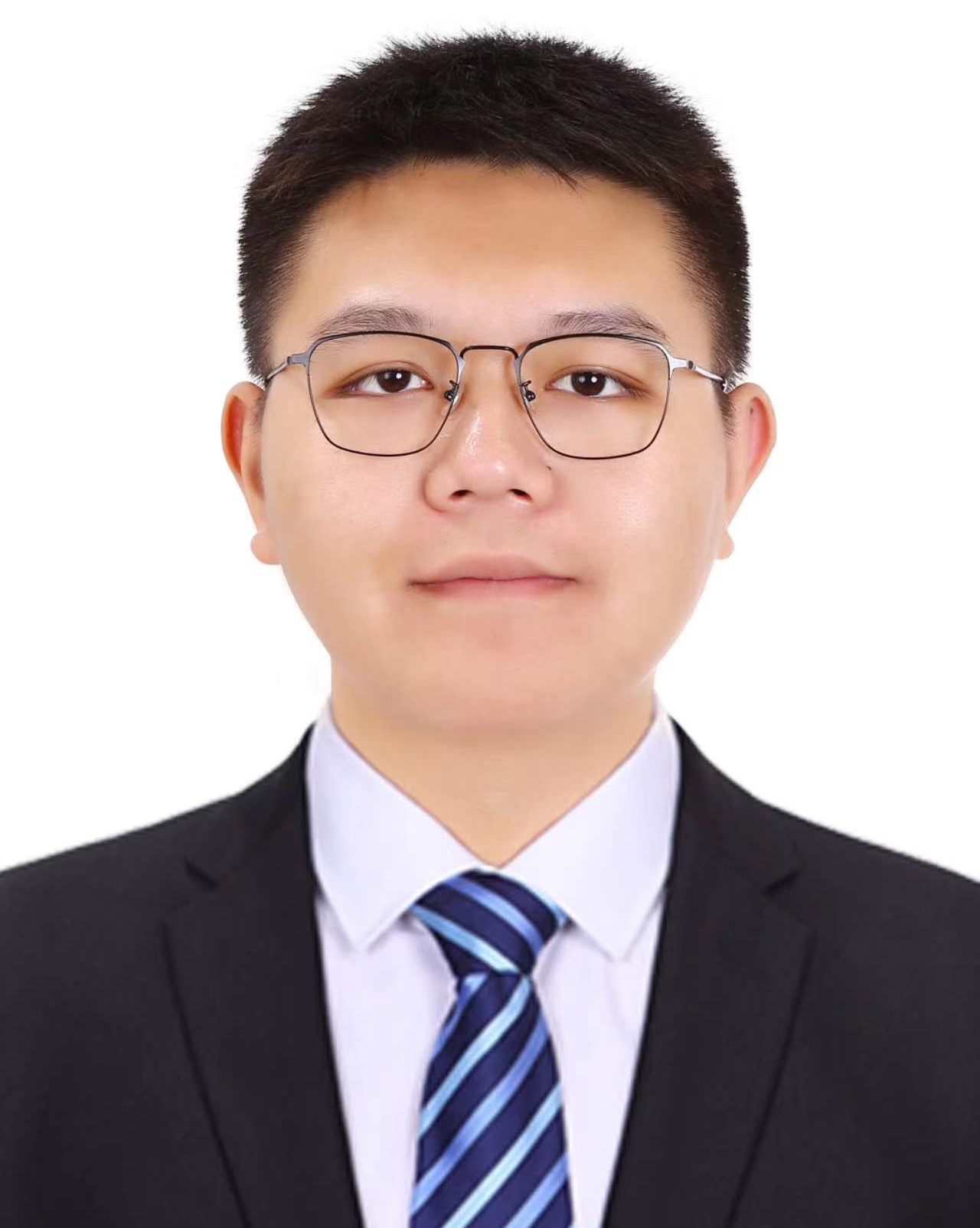}}]{Shiji Zhao} received his B.S. degree in the School of Computer Science and Engineering, Beihang University (BUAA), China. He is now a Ph.D student in the Institute of Artificial Intelligence, Beihang University (BUAA), China. His research interests include computer vision, deep learning and adversarial robustness in machine learning.
\end{IEEEbiography}
% \vfill

\end{document}